\pdfoutput=1

\documentclass[11pt]{article}

\usepackage{acl}

\usepackage{times}
\usepackage{latexsym}

\usepackage[T1]{fontenc}

\usepackage[utf8]{inputenc}

\usepackage{microtype}
\usepackage{times}
\usepackage{latexsym}
\usepackage[smallerops]{newtxmath}
\usepackage{amsmath}

\usepackage{amssymb}
\usepackage{booktabs}
\usepackage{enumitem}
\usepackage{algorithm}
\usepackage{algpseudocode}
\usepackage{multicol}
\usepackage{multirow}
\usepackage{makecell}
\usepackage{supertabular}
\usepackage{xspace}
\usepackage{bm}
\usepackage{IEEEtrantools}
\usepackage{graphicx}
\usepackage{xcolor}
\usepackage{tikz}
\usetikzlibrary{bayesnet}
\usepackage{caption}
\usepackage{subcaption}
\usepackage{comment}

\title{Distributed Marker Representation for Ambiguous \\ Discourse Markers and Entangled Relations}

\author{
\parbox{\linewidth}{
\centering
Dongyu Ru$^1$, Lin Qiu$^1$, Xipeng Qiu$^2$, Yue Zhang$^3$, Zheng Zhang$^1$
}\\
$^1$Amazon AWS AI ~~~~$^2$School of Computer Science, Fudan University \\
$^3$School of Engineering, Westlake University\\
{\texttt{\{rudongyu,quln,zhaz\}@amazon.com}} \\
{\texttt{xpqiu@fudan.edu.cn}}\\
{\texttt{zhangyue@westlake.edu.cn}}
}

\begin{document}
\maketitle
\begin{abstract}
Discourse analysis is an important task because it models intrinsic semantic structures between sentences in a document. Discourse markers are natural representations of discourse in our daily language. One challenge is that the markers as well as pre-defined and human-labeled discourse relations can be ambiguous when describing the semantics between sentences. We believe that a better approach is to use a contextual-dependent distribution over the markers to express discourse information. In this work, we propose to learn a Distributed Marker Representation (DMR) by utilizing the (potentially) unlimited discourse marker data with a latent discourse sense, thereby bridging markers with sentence pairs. Such representations can be learned automatically from data without supervision, and in turn provide insights into the data itself. Experiments show the SOTA performance of our DMR on the implicit discourse relation recognition task and strong interpretability. Our method also offers a valuable tool to understand complex ambiguity and entanglement among discourse markers and manually defined discourse relations.

\end{abstract}

\section{Introduction}
\label{sec:intro}
Discourse analysis is a fundamental problem in natural language processing. It studies the linguistic structures beyond the sentence boundary and is a component of chains of thinking. Such structural information has been widely applied in many downstream applications, including information extraction~\cite{peng-etal-2017-cross}, long documents summarization~\cite{cohan-etal-2018-discourse}, document-level machine translation~\cite{chen-etal-2020-modeling}, conversational machine reading~\cite{gao-etal-2020-discern}.

\begin{figure}
    \centering
    \includegraphics[width=0.8\columnwidth]{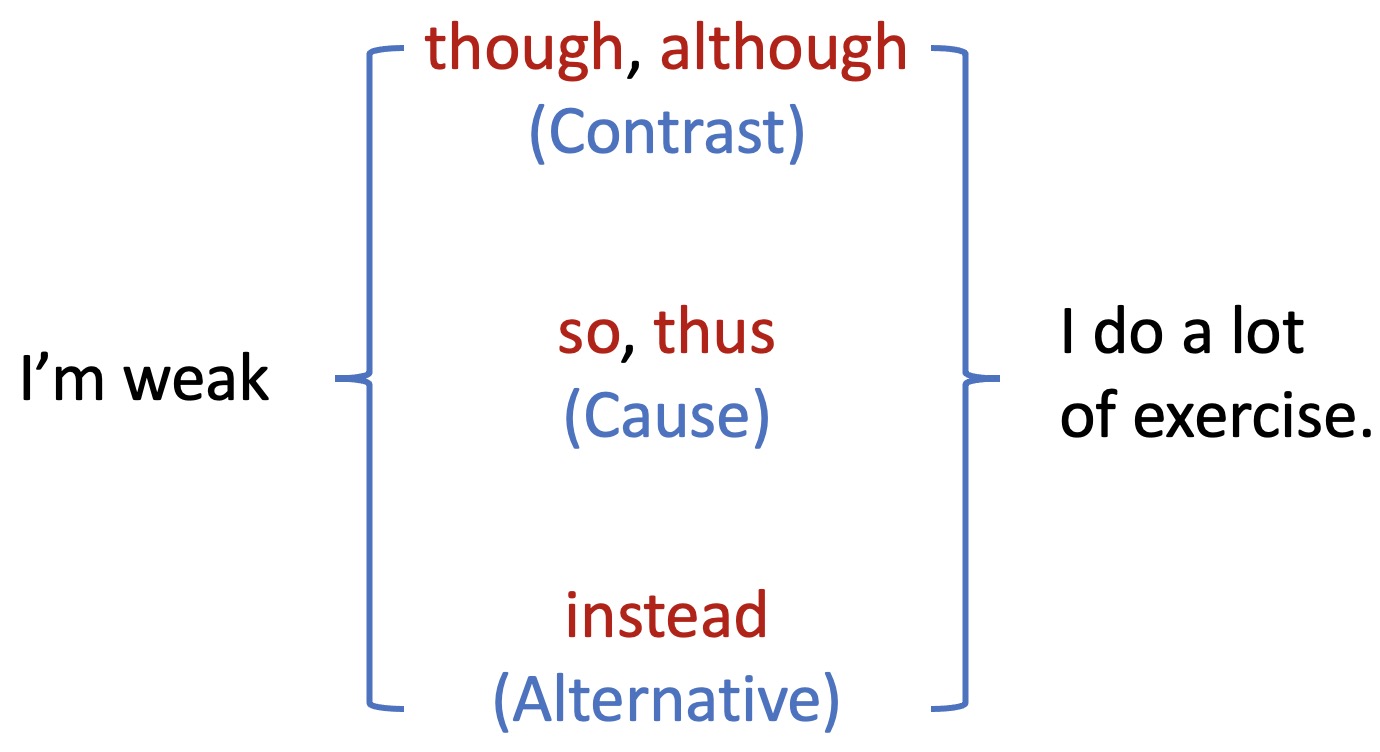}
    \caption{Entangled discourse relations and corresponding markers between clauses. As shown in the figure, there exist diverse discourse relations (marked in blue) and corresponding markers (marked in red) for the same pair of clauses. It suggests that the semantic meaning of different discourse relations can be entangled to each other.}
    \label{fig:head-example}
\end{figure}

Discourse relation recognition (DRR) focuses on semantic relations, namely, discourse senses between sentences or clauses. Such inter-sentence structures are sometimes explicitly expressed in natural language by discourse connectives, or markers (e.g., \emph{and}, \emph{but}, \emph{or}). The availability of these markers makes it easier to identify corresponding relations~\cite{pitler2008easily}, as is in the task of explicit discourse relation recognition (EDRR), since there is strong correlation between discourse markers and relations. On the contrary, implicit discourse relation recognition (IDRR), where markers are missing, remains a more challenging problem. 

Prior work aims to address such challenges by making use of discourse marker information over explicit data in learning implicit discourse relations, either by injecting marker prediction knowledge into a representation model \cite{zhou-etal-2010-predicting, braud-denis-2016-learning}, or transferring the marker prediction task into implicit discourse relation prediction by manually defining a marker-relation mapping \cite{xiang-etal-2022-connprompt, zhou2022prompt}. It has been shown that discourse marker information can effectively improve relation prediction results. Nevertheless, relatively little work has investigated various subtleties concerning the correlation between discourse markers and discourse relations, and their effect to IDRR in further detail.



To properly model discourse relations and markers, we need to consider that manually-defined discourse relations can be semantically entangled and markers are ambiguous. As shown in Fig.~\ref{fig:head-example}, for a pair of clauses, based on different emphasis on semantics, we have different choices on discourse relations and their corresponding markers. The existence of multiple plausible discourse relations indicates the entanglement between their semantic meaning. Besides, discourse markers and relations do not exclusively map to each other. As an example, ``Ann went to the movies, \textbf{and} Bill went home'' (\emph{Temporal.Synchrony}) and ``Ann went to the movies, \textbf{and} Bill got upset'' (\emph{Contingency.Cause}) both use the marker \textbf{and} but express different meanings.
Identifying relations based on single markers are difficult in certain scenarios because of such ambiguity.
Thus, a discrete and deterministic mapping between discourse relations and markers can not precisely express the correlations between them.

Based on the study of above issues, we propose to use \textbf{D}istributed \textbf{M}arker \textbf{R}epresentation to enhance the informativeness of discourse expression. Specifically, We use a probabilistic distribution on markers or corresponding latent senses instead of a single marker or relation to express discourse semantics.
We introduce a bottleneck in the latent space, namely a discrete latent variable indicating discourse senses, to capture semantics between clauses. The latent sense then produces a \emph{distribution} of plausible markers to reflect its surface form. This probabilistic model, which we call DMR, naturally deals with ambiguities between markers and entanglement among the relations.
We show that the latent space reveals a hierarchical marker-sense clustering, and that entanglement among relations are currently under-reported. Empirical results on the IDRR benchmark Penn Discourse Tree Bank 2 (PDTB2)~\cite{prasad2008penn} shows the effectiveness of our framework.
We summarize our contributions as follows:
\begin{itemize}[leftmargin=1pt, itemindent=1pc]
\item We propose a latent-space learning framework for discourse relations and effectively optimize it with cheap marker data.\footnote{Code is publicly available at: \url{https://github.com/rudongyu/DistMarker}}
\item With the latent bottleneck and corresponding probabilistic modeling, our framework achieves the SOTA performance on implicit discourse relation recognition without a complicated architecture design.
\item We investigate the ambiguity of discourse markers and entanglement among discourse relations to explain the plausibility of probabilistic modeling of discourse relations and markers. 
\end{itemize}

\section{Related Work}
\label{sec:related}
Discourse analysis~\cite{brown1983discourse,joty2019discourse,mccarthy2019discourse},
targets the discourse relation between adjacent sentences. It has attracted attention beyond intra-sentence semantics. It is formulated into two main tasks: explicit discourse relation recognition and implicit discourse relation recognition, referring to the relation identification between a pair of sentences with markers explicitly included or not. While EDRR has achieved satisfactory performance~\cite{pitler2008easily} with wide applications, IDRR remains to be challenging~\cite{pitler2009automatic,zhang2015shallow,rutherford2017systematic,shi2019next}. Our work builds upon the correlation between the two critical elements in discourse analysis: discourse relations and markers.

Discourse markers have been used for not only marker prediction training~\cite{malmi2018automatic}, but also for improving the performance of IDRR~\cite{marcu2002unsupervised,rutherford-xue-2015-improving} and representation learning~\cite{jernite2017discourse}. Prior efforts on exploring markers have found that training with discourse markers can alleviate the difficulty on IDRC~\cite{sporleder_lascarides_2008,zhou-etal-2010-predicting,braud-denis-2016-learning}. Compared to their work, we focus on a unified representation using distributed markers instead of relying on transferring from explicit markers to implicit relations. \citet{jernite2017discourse} first extended the usage of markers to sentence representation learning, followed by~\citet{nie2019dissent,sileo2019mining} which introduced principled pretraining frameworks and large-scale marker data. \citet{xiang-etal-2022-connprompt, zhou2022prompt} explored the possibility of connecting markers and relations with prompts. In this work, we continue the line of improving the expression of discourse information as distributed markers.

\section{Distributed Marker Representation Learning}
\label{sec:method}

We elaborate on the probabilistic model in Sec.~\ref{sec:method:overview} and its implementation with neural networks in Sec.~\ref{sec:method:model}. We then describe the way we optimize the model (Sec.~\ref{sec:method:opt}).

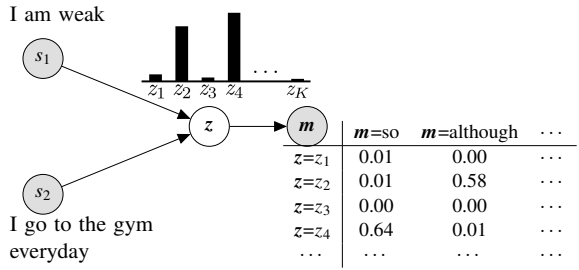
\begin{figure}[t]
\centering
\scalebox{0.75}{
\tikz{
 \node[latent] (z) at (1, 0) {$\bm{z}$}; %
 \node[obs, left=of z, xshift=-1.2cm, yshift=1.2cm] (x) {$s_1$};%
 \node[obs, left=of z, xshift=-1.2cm, yshift=-1.2cm] (y) {$s_2$};%
 \node[obs, right=of z] (c) {$\bm{m}$};
 \node[text width=3cm] at (-1, 2) {I am weak};
 \node[text width=3cm] at (-1, -2) {I go to the gym everyday};
 \node [above=of z, yshift=-1cm, xshift=0.3cm] {
    \includegraphics[width=3cm]{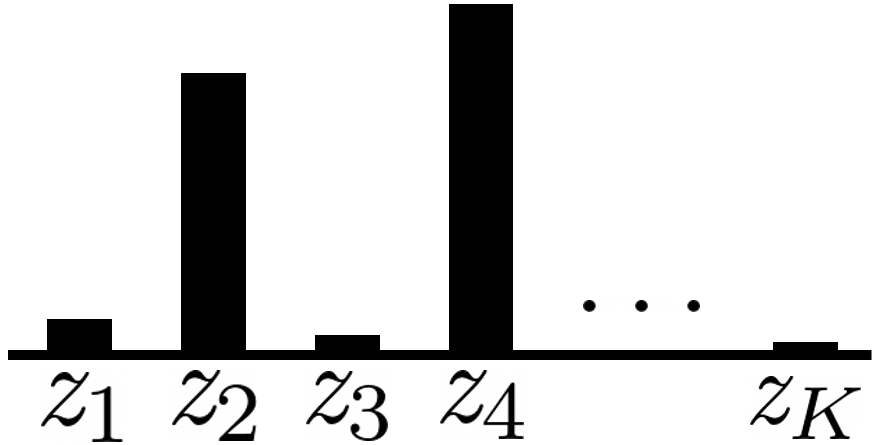}
 };
 \node [right=of c, xshift=-2cm, yshift=-1.2cm] {
    \scalebox{0.9}{
    \begin{tabular}{c|ccc}
        & $\bm{m}$=so & $\bm{m}$=although & $\cdots$ \\\hline
        $\bm{z}$=$z_1$ & 0.01 & 0.00 & $\cdots$ \\
        $\bm{z}$=$z_2$ & 0.01 & 0.58 & $\cdots$ \\
        $\bm{z}$=$z_3$ & 0.00 & 0.00 & $\cdots$ \\
        $\bm{z}$=$z_4$ & 0.64 & 0.01 & $\cdots$ \\
        $\cdots$ & $\cdots$ & $\cdots$ & $\cdots$
    \end{tabular}}
 };
 \edge {z} {c};
 \edge {x, y} {z};
}
}
\caption{The graphical model of $p(\bm{m}|s_1, s_2)$. $\bm{z}$ is the latent variable indicating the latent sense, namely the semantic relation between two clauses. $K$ is the number of candidate values for the random variable $\bm{z}$.}
\label{fig:gm}
\vspace{-10pt}
\end{figure}


\subsection{Probabilistic Formulation}
\label{sec:method:overview}
We learn the distributed marker representation by predicting markers given pairs of sentences.
We model the distribution of markers by introducing an extra latent variable $\bm{z}$ which indicates the latent senses between two sentences.
We assume the distribution of markers depends only on the latent senses, and is independent of the original sentence pairs when $\bm{z}$ is given, namely $\bm{m} \perp (s_1, s_2) | \bm{z}$. 
\begin{IEEEeqnarray}{c}
   p_{\psi, \phi}(\bm{m}|s_1, s_2) = \sum_{\bm{z}} p_{\psi}(\bm{z}|s_1, s_2) \cdot p_{\phi}(\bm{m}|\bm{z}),
   \label{eq:prob-model}
\end{IEEEeqnarray}
where the latent semantic senses $\bm{z}$ describes the unambiguous semantic meaning of $m$ in the specific context, and our target is to model the probabilistic distribution $p(\bm{m}|s_1, s_2)$ with $\bm{z}$. The probabilistic model is depicted in Fig.~\ref{fig:gm} with an example.

The key inductive bias here is that we assume the distribution of discourse markers is independent of the original sentence pairs given the latent semantic senses (Eq.~\ref{eq:prob-model}). This formulation is based on the intuition that humans decide the relationship between two sentences in their cognitive worlds first, then pick one proper expression with a mapping from latent senses to expressions (which we call \emph{z2m} mapping in this paper) without reconsidering the semantic of sentences. Decoupling the \emph{z2m} mapping from the distribution of discourse marker prediction makes the model exhibit more interpretability and transparency.

Therefore, the probabilistic distribution of $p_{\psi, \phi}(\bm{m}|s_1, s_2)$ can be decomposed into $p_{\psi}(m|\bm{z})$ and $p_{\phi}(\bm{z}|s_1, s_2)$ based on the independence assumption above. $\psi$ and $\phi$ denote parameters for each part \footnote{We omit the subscript of parameters $\psi$ and $\phi$ in some expressions later for conciseness.}. 
The training objective with latent senses included is to maximize the likelihood on large-scale corpus under this assumption:
\begin{IEEEeqnarray}{c}
   \mathcal{L}(\psi, \phi) = \mathbb{E}_{(s_1, s_2, m)\sim D} \log p_{\psi, \phi}(m|s_1, s_2).
   \label{eq:likelihood}
\end{IEEEeqnarray}


\subsection{Neural Architecture}
\label{sec:method:model}

Our model begins by processing each sentence with an encoder \texttt{SentEnc}:
\begin{IEEEeqnarray}{c}
h = \texttt{SentEnc}_{\psi_s}([s_1, \texttt{[SEP]}, s_2]),
\end{IEEEeqnarray}
where $h\in \mathbb{R}^d$ denote the sentence pair representation in $d$ dimensions for $s_1$ and $s_2$. $\psi_s$ are parameters of the sentence encoder. The encoder is instantiated as a pre-trained language model in practice.

\begin{algorithm*}[t]
  \caption{EM Optimization for Discourse Marker Training with Latent Senses}\label{alg:em}
  \begin{algorithmic}[1]
    \State Initialize model parameters as $\psi^0, \phi^0$.
    \While {not converge} \Comment{$t$-th iteration}
        \State Sample a batch of examples for EM optimization.
        \For {each example $(s_1, s_2, m)$ in the EM batch}
        \State Calculate and save the posterior $p(\bm{z}|s_1, s_2, m)$ according to $\psi^{(t)}, \phi^{(t)}$.
        \EndFor
        \For {each example $(s_1, s_2, m)$ in the EM batch}
        \State Estimate $\mathbb{E}_{p(z|s_1,s_2,m)}[\log p_{\psi, \phi}(m, z|s_1, s_2)]$ according to $\psi^{(t)}, \phi^{(t)}$.
        \Comment{E-step}
        \EndFor
        \State Update parameters $\psi$ to $\psi^{(t+1)}$ in mini-batch with the gradient calculated as $\nabla_\psi\mathcal{L}(\psi, \phi^{(t)})$.
        \State Update parameters $\phi$ to $\phi^{(t+1)}$ according to the updated $\psi^{(t+1)}$ and the gradient $\nabla_\phi\mathcal{L}(\psi^{(t+1)}, \phi)$.
        \Comment{M-step}
    \EndWhile
  \end{algorithmic}
  \label{alg:optimization}
\end{algorithm*}

Then we use two linear layers to map the pair representation $h$ to the distribution of $\bm{z}$ as below:
\begin{IEEEeqnarray}{c}
h_z = \psi_{w1}\cdot h + \psi_{b1}, \\
p_\theta(z|s_1, s_2) = \texttt{softmax} (\psi_{w2} \cdot h_z + \psi_{b2}),
\label{eq:softmax}
\end{IEEEeqnarray}
where $\psi_{w1}\in\mathbb{R}^{d\times4d}, \psi_{b1}\in\mathbb{R}^d, \psi_{w2}\in\mathbb{R}^{K\times d}, \psi_{b2}\in\mathbb{R}^K$ are trainable parameters. $K$ is the dimension of latent discourse senses.

The parameter $\psi_{w2}$ not only acts as the mapping from representation $h_z$ to $\bm{z}$'s distribution, but can also be seen as an embedding lookup table for the $K$ values of $\bm{z}$. Each row in $\psi_{w2}$ is a representation vector for the corresponding value, as an anchor in the companion continuous space of $\bm{z}$.

To parameterize the \emph{z2m} mapping, the parameter $\phi\in\mathbb{R}^{K\times N}$ is defined as a probabilistic transition matrix from latent semantic senses $z$ to markers $m$ (in log space), where $N$ is the number of candidate markers:
\begin{IEEEeqnarray}{c}
\log p_\phi(m|z) = \log \texttt{softmax} (\phi),
\end{IEEEeqnarray}
where $\psi=(\psi_s, \psi_{w1}, \psi_{b1}, \psi_{w2}, \psi_{b2}), \phi$ are the learnable parameters for parameterize the distribution $p_{\psi, \phi} (\bm{m} | s_1, s_2)$.

\subsection{Optimization}
\label{sec:method:opt}

We optimize the parameters $\psi$ and $\phi$ with the classic EM algorithm due to the existence of the latent variable $\bm{z}$.
The latent variable $\bm{z}$ serves as a regularizer during model training.
In the E-step of each iteration, we obtain the posterior distribution $p(z|s_1, s_2, m)$ according to the parameters in the current iteration $\psi^{(t)}, \phi^{(t)}$ as shown in Eq.~\ref{eq:estep}.

Based on our assumption that $\bm{m} \perp (s_1, s_2) | \bm{z}$, we can get the posterior distribution:
\begin{IEEEeqnarray}{rl}
p(z|s_1, s_2, m) &= \frac{p(m | s_1, s_2, z)\cdot p(z|s_1, s_2)}{p(m|s_1, s_2)} \nonumber\\
&= \frac{p(m | z)\cdot p(z|s_1, s_2)}{p(m|s_1, s_2)} \nonumber\\
& \propto p_{\psi^{(t)}}(z|s_1, s_2) \cdot p_{\phi^{(t)}}(m|z). \label{eq:estep}
\end{IEEEeqnarray}


In M-step, we optimize the parameters $\psi, \phi$ by maximizing the expectation of joint log likelihood on estimated posterior $p(z|s_1,s_2,m)$. The updated parameters $\psi^{(t+1)}, \phi^{(t+1)}$ for the next iteration can be obtained as in Eq.~\ref{eq:mstep}.
\begin{IEEEeqnarray}{rl}
\label{eq:mstep}
\psi^{(t+1)}&, \phi^{(t+1)}= \\ \nonumber &\underset{\psi,\phi}{\arg\max}\mathbb{E}_{p(z|s_1,s_2,m)}[\log p_{\psi, \phi}(m, z|s_1, s_2)].
\end{IEEEeqnarray}

In practice, the alternative EM optimization can be costly and unstable due to the expensive expectation computation and the subtlety on hyperparameters when optimizing $\psi$ and $\phi$ jointly.
We alleviate the training difficulty by empirically estimating the expectation on mini-batch and separate the optimization of $\psi$ and $\phi$.
We formulate the loss functions as below, for separate gradient descent optimization of $\psi$ and $\phi$:

{\small
\begin{IEEEeqnarray*}{c}
\mathcal{L}(\psi, \phi^{(t)}) = \text{KLDiv}(p(z|s_1, s_2, m), p_{\psi, \phi^{(t)}}(m, z|s_1, s_2)), \\
\mathcal{L}(\psi^{(t+1)}, \phi) = -\log p_{\psi^{(t+1)},\phi}(m|s_1, s_2),
\end{IEEEeqnarray*}
}

where $\phi^{(t)}$ means the value of $\phi$ before the $t$-th iteration and $\psi^{(t+1)}$ means the value of $\psi$ after the $t$-th iteration of optimization. $\text{KLDiv}$ denotes the Kullback-Leibler divergence. The overall optimization algorithm is summarized in Algorithm~\ref{alg:em}.



\section{Experiments}
\label{sec:exp}
DMR adopts a latent bottleneck for the space of latent discourse senses. We first prove the effectiveness of the latent variable and compare against current SOTA solutions on the IDRR task. We then examine what the latent bottleneck learned during training and how it addresses the ambiguity and entanglement of discourse markers and relations.

\subsection{Dataset}
We use two datasets for learning our DMR model and evaluating its strength on downstream implicit discourse relation recognition, respectively. See Appendix~\ref{supp:implementation} for statistics of the datasets.

\paragraph{Discovery Dataset}~\cite{sileo2019mining}
is a large-scale discourse marker dataset extracted from commoncrawl web data, the Depcc corpus~\cite{panchenko-etal-2018-building}. It contains 1.74 million sentence pairs with a total of 174 types of explicit discourse markers between them. Markers are automatically extracted based on part-of-speech tagging. We use top-k accuracy \textbf{ACC@k} to evaluate the marker prediction performance on this dataaset.
Note that we use explicit markers to train DMR but evaluate it on IDRR thanks to different degrees of verbosity when using markers in everyday language.

\begin{table*}[t]
    \centering
    \scalebox{0.9}{
    \begin{tabular}{cccc}
    \toprule
        \textbf{Model} & \textbf{Backbone} & \textbf{macro-F$_1$} & \textbf{ACC} \\\midrule
        IDRR-C\&E \cite{dai2019regularization} & ELMo & 33.41 & 48.23 \\
        MTL-MLoss \cite{nguyen-etal-2019-employing} & ELMo & - & 49.95 \\
        BERT-FT \cite{kishimoto2020adapting} & BERT & - & 54.32 \\
        HierMTN-CRF \cite{wu2020hierarchical} & BERT & 33.91 & 52.34 \\
        BMGF-RoBERTa \cite{liu2021importance} & RoBERTa & - & 58.13 \\
        MTL-MLoss-RoBERTa$^\dagger$ \cite{nguyen-etal-2019-employing} & RoBERTa & 38.10 & 57.72 \\
        HierMTN-CRF-RoBERTa$^\dagger$ \cite{wu2020hierarchical} & RoBERTa& 38.28 & 58.61 \\
        LDSGM \cite{wu2022label} & RoBERTa & 40.49 & 60.33 \\
        PCP-base \cite{zhou2022prompt} & RoBERTa &
        41.55 & 60.54 \\
        PCP-large \cite{zhou2022prompt} & RoBERTa &
        \textbf{44.04} & 61.41 \\
        \midrule
        DMR-base$_{\text{w/o \emph{z}}}$ & RoBERTa & 37.24 & 59.89 \\
        DMR-large$_{\text{w/o \emph{z}}}$ & RoBERTa & 41.59 & 62.35 \\
        DMR-base & RoBERTa & 42.41 & 61.35 \\
        DMR-large & RoBERTa & 43.78 & \textbf{64.12} \\
    \bottomrule
    \end{tabular}
    }
    \caption{Experimental Results of Implicit Discourse Relation Classification on PDTB2. Results with $\dagger$ are from \citet{wu2022label}. DMR-large and DMR-base adopt roberta-large and roberta-base as \texttt{SentEnc}, respectively.}
    \label{tab:discourse_relation_comp}
\vspace{-10pt}
\end{table*}

\paragraph{Penn Discourse Tree Bank 2.0 (PDTB2)}~\cite{prasad2008penn} is a popular discourse analysis benchmark with manually-annotated discourse relations and markers on Wall Street Journal articles. We perform the evaluation on its implicit part with 11 major second-level relations included.
We follow~\cite{ji2015one} for data split, which is widely used in recent studies for IDRR. \textbf{Macro-F$_1$} and \textbf{ACC} are metrics for IDRR performance. We note that although annotators are allowed to annotate multiple senses (relations), only 2.3\% of the data have more than one relation. Therefore whether DMR can capture more entanglement among relations is of interest as well (Sec.~\ref{sec:exp:discussion}).

\subsection{Baselines}
We compare our DMR model with competitive baseline approaches to validate the effectiveness of DMR. For the IDRR task, we compare DMR-based classifier with current SOTA methods, including BMGF~\cite{liu2021importance}, which combines representation, matching, and fusion; LDSGM~\cite{wu2022label}, which considers the hierarchical dependency among labels; the prompt-based connective prediction method, PCP~\cite{zhou2022prompt} and so on. 
For further analysis on DMR, we also include a vanilla sentence encoder without the latent bottleneck as an extra baseline, denoted as BASE.

\subsection{Implementation Details}

Our DMR model is trained on 1.57 million examples with 174 types of markers in Discovery dataset. We use pretrained RoBERTa model~\cite{liu2019roberta} as \texttt{SentEnc} in DMR. We set the default latent dimension $K$ to 30.
More details regarding the implementation of DMR can be found in Appendix~\ref{supp:implementation}.


For the IDRR task, we strip the marker generation part from the DMR model and use the hidden state $h_z$ as the pair representation.
$\text{BASE}$ uses the \texttt{[CLS]} token representation as the representation of input pairs.
A linear classification layer is stacked on top of models to predict relations.

\subsection{Implicit Discourse Relation Recognition}

\begin{table}[]
\scalebox{0.85}{
\begin{tabular}{cccc}
\toprule
      & BMGF  & LDSGM  & DMR  \\ \midrule
Comp.Concession  & 0. & 0. & 0. \\
Comp.Contrast  & 59.75 & \textbf{63.52}  & 63.16 \\
Cont.Cause & 59.60  & \textbf{64.36}  & 62.65  \\
Cont.Pragmatic Cause  & 0. & 0. & 0. \\
Expa.Alternative  & 60.0  & \textbf{63.46}  & 55.17 \\
Expa.Conjunction  & \textbf{60.17} & 57.91  & 58.54 \\
Expa.Instantiation & 67.96 & \textbf{72.60}  & 72.16 \\
Expa.List & 0. & 8.98 & \textbf{36.36} \\
Expa.Restatement & 53.83 & 58.06 & \textbf{59.19} \\
Temp.Async       & 56.18 & 56.47 & \textbf{59.26} \\
Temp.Sync        & 0.    & 0.    & 0.    \\
Macro-f1         & 37.95 & 40.49 & \textbf{42.41} \\ \bottomrule
\end{tabular}
}
\caption{Experimental Results of Implicit Discourse Relation Recognition on PDTB2 Second-level Senses}
    \label{tab:discourse_fine}
\end{table}

We first validate the effectiveness of modeling latent senses on the challenging IDRR task.

\paragraph{Main Results}
DMR demonstrates comparable performance with current SOTAs on IDRR, but with a simpler architecture. As shown in Table~\ref{tab:discourse_relation_comp}, DMR leads in terms of accuracy by 2.7pt and is a close second in macro-F$_1$.

The results exhibit the strength of DMR by more straightforwardly modeling the correlation between discourse markers and relations. Despite the absence of supervision on discourse relations during DMR learning, the semantics of latent senses distilled by EM optimization successfully transferred to manually-defined relations in IDRR.

Based on the comparison to DMR without latent \emph{z}, we observe a significant performance drop resulted from the missing latent bottleneck. It indicates that the latent bottleneck in DMR serves as a regularizer to avoid overfitting on similar markers.

\begin{figure}
    \centering
    \includegraphics[width=0.8\linewidth]{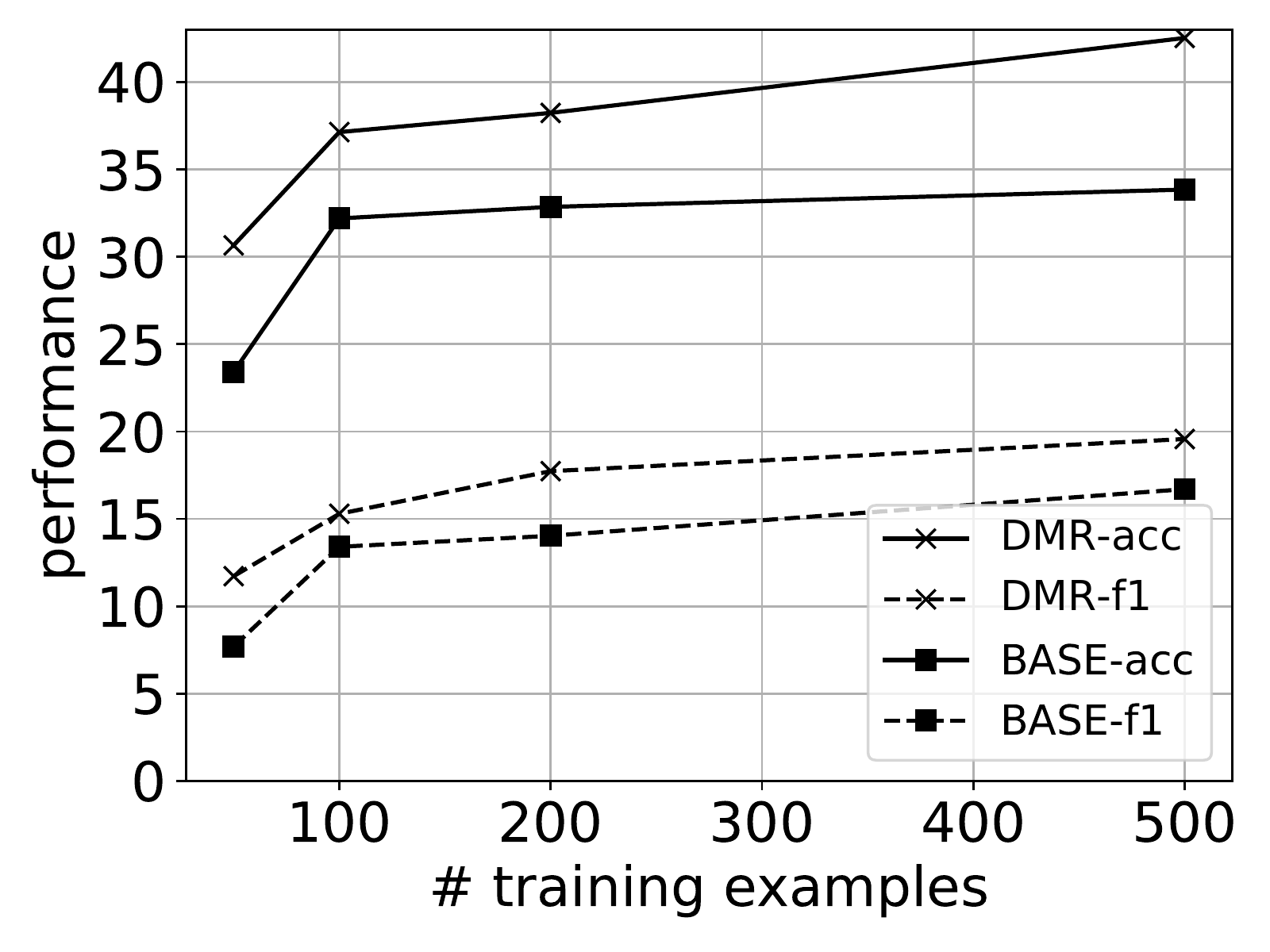}
    \caption{Few-shot IDRR Results on PDTB2}
    \label{fig:discourse}
\end{figure}

\begin{table}[t]
\centering
\scalebox{0.8}{
\begin{tabular}{cccccccc}
\toprule
    \multicolumn{2}{c}{\# Training Examples} & 25    &  100   & 500   & full (10K)  \\ \midrule
    \multirow{2}{*}{BASE} & ACC & -  & 32.20 & 33.85 & 59.45 \\ 
     & F1  & -  &  13.40 & 16.70 & 34.34 \\ \hline
    \multirow{2}{*}{BASE$_p$} & ACC &    -   &   33.76 &   37.56   & 60.90      \\
    & F1  &    -   &    13.54 & 17.21  &   35.45    \\ \hline
    \multirow{2}{*}{BASE$_g$} & ACC & 19.12 & 34.07 &  39.23 & \textbf{63.19} \\
     & F1  & 5.75  & 13.72 & 19.27 & 36.59 \\ \hline
     \multirow{2}{*}{DMR} & ACC & \textbf{21.32} & \textbf{37.14} & \textbf{42.53} & 62.97 \\
    & F1  & \textbf{7.01}  & \textbf{15.29} & \textbf{19.57} & \textbf{39.33} \\
     \bottomrule
\end{tabular}
}
\caption{Few-shot IDRR Results on PDTB2}
\label{tab:discourse_relation}
\end{table}

\paragraph{Fine-grained Performance}
We list the fine-granined performance of DMR and compare it with SOTA approaches on second-level senses of PDTB2. As shown in Table~\ref{tab:discourse_fine}, DMR achieves significant improvements on relations with little supervision, like \emph{Expa.List} and \emph{Temp.Async}.
The performance of majority classes, e.g. \emph{Expa.Conjunction}, are slightly worse. It may be caused by the entanglement between \emph{Expa.Conjunction} and \emph{Expa.List} to be discussed in Sec.~\ref{sec:exp:discussion}. In summary, DMR achieves better overall performance by maintaining equilibrium among entangled relations with different strength of supervision.

\paragraph{Few-shot Analysis}

Fig.~\ref{fig:discourse} shows DMR achieves significant gains against BASE in few-shot learning experiments. The results are averaged on 3 independent runs for each setting. In fact, with only $\sim$60\% of annotated data, DMR achieves the same performance as BASE with full data by utilizing the cheap marker data more effectively.

To understand the ceiling of the family of such BERT-based pretrained model with markers as an extra input, we augment the data in two ways: $\text{BASE}_g$ inserts the groundtruth marker, and $\text{BASE}_p$ where the markers are predicted by a model\footnote{They also use the RobERTa model as a backbone.} officially released by Discovery~\cite{sileo2019mining}.
Table~\ref{tab:discourse_relation} presents the results where the informative markers are inserted to improve the performance of BASE, following the observations and ideas from ~\cite{zhou-etal-2010-predicting, pitler2008easily}. DMR continues to enjoy the lead, even when the markers are groundtruth (i.e. $\text{BASE}_g$), suggesting DMR's hidden state contains more information than single markers.

\subsection{Analysis \& Discussion}
\label{sec:exp:discussion}

\begin{table}[t]
    \centering
    \scalebox{0.8}{
    \begin{tabular}{ccccc}
    \toprule
        \textbf{Model} & \textbf{ACC@1} & \textbf{ACC@3} & \textbf{ACC@5} & \textbf{ACC@10} \\\midrule
        Discovery & \textbf{24.26}& \textbf{40.94}& 49.56& 61.81\\
        DMR$_{30}$ & 8.49 & 22.76 & 33.54 & 48.11 \\
        DMR$_{174}$ & 22.43 & 40.92 & \textbf{50.18} & \textbf{63.21} \\
    \bottomrule
    \end{tabular}
    }
    \caption{Experimental results of marker prediction on the Discovery test set. DMR$_{30}$ and DMR$_{174}$ indicate the models with the dimension K equals to 30 and 174 respectively.}
    \label{tab:marker-prediction}
\vspace{-10pt}
\end{table}


\begin{table}[t]
    \centering
    \scalebox{0.8}{
    \begin{tabular}{ccc}
    \toprule
        \textbf{Marker} & \textbf{1st Cluster} & \textbf{2nd Cluster} \\\midrule
        additionally & $\bm{z}_1$:~~
        \makecell[c]{as a result,\\in turn, \\simultaneously} & $\bm{z}_{20}$:~~
        \makecell[c]{for example, \\ for instance, \\specifically} \\ \midrule
        amazingly & $\bm{z}_{9}$:
        \makecell[c]{thankfully,\\ fortunately, \\luckily} & $\bm{z}_{21}$:
        \makecell[c]{oddly, \\ strangely, \\unfortunately} \\ \midrule
        but & $\bm{z}_{19}$:
        \makecell[c]{indeed,\\ nonetheless, \\nevertheless} & $\bm{z}_{24}$:
        \makecell[c]{anyway, \\and,\\ well} \\
    \bottomrule
    \end{tabular}
    }
    \caption{Top 2 clusters of three random sampled markers. Each cluster corresponds to a latent $\bm{z}$ coupled with its top 3 markers.}
    \label{tab:m2z}
\end{table}

\paragraph{Marker Prediction}
 The performance of DMR on marker prediction is sensitive to the capacity of the bottleneck. When setting $K$ to be the number of markers (174), it matches and even outperforms the Discovery model which directly predicts the markers on the same data (Table~\ref{tab:marker-prediction}). A smaller $K$ sacrifices marker prediction performance but it can cluster related senses, resulting in more informative and interpretable representation.

Multiple markers may share similar meanings when connecting sentences. Thus, evaluating the performance of marker prediction simply on top1 accuracy is inappropriate. In Table~\ref{tab:marker-prediction}, we demonstrated the results on ACC@k and observed that DMR(K=174) gets better performance against the model optimized by an MLE objective when k gets larger. We assume that it comes from the marker ambiguity. Our DMR models the ambiguity better, thus with any of the plausible markers easier to be observed in a larger range of predictions but more difficult as top1. To prove the marker ambiguity more directly, we randomly sample 50 examples to analyze their top5 predictions. The statistics show that over 80\% of those predictions have plausible explanations. To conclude, considerable examples have multiple plausible markers thus ACC@k with larger k can better reflect the true performance on marker prediction, where DMR can beat the MLE-optimized model.


\paragraph{\emph{z2m} Mapping}

\begin{figure}[t]
    \centering
    \begin{subfigure}[t]{0.8\columnwidth}
         \centering
         \includegraphics[width=\textwidth]{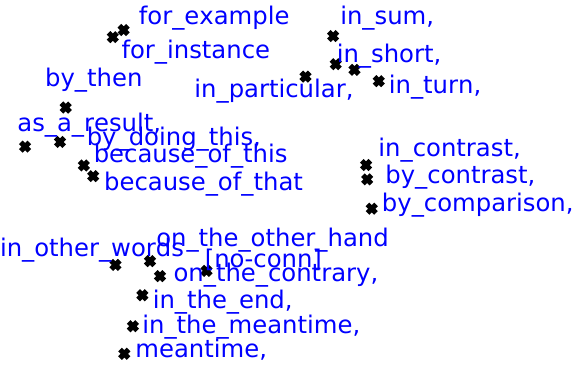}
         \caption{The cropped T-SNE visualization of discourse markers from the BASE PLM.}
         \label{fig:plm-tsne-crop}
    \end{subfigure}
    \begin{subfigure}[t]{0.8\columnwidth}
         \centering
         \includegraphics[width=\textwidth]{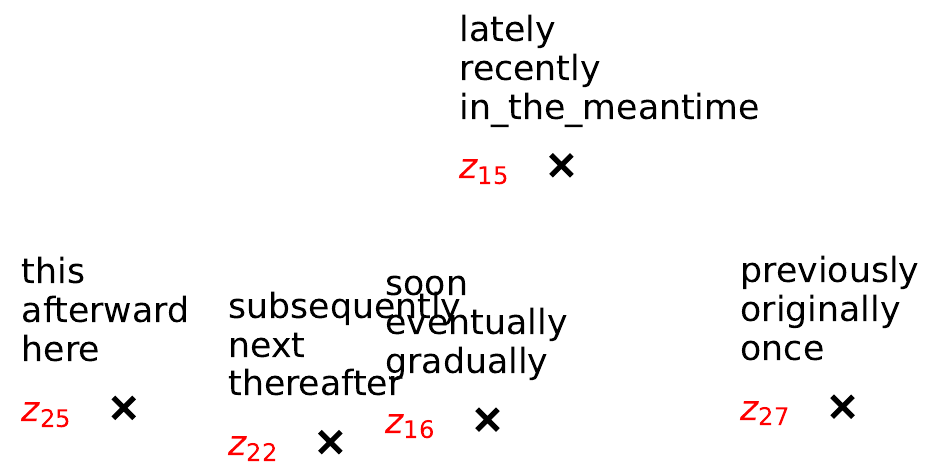}
         \caption{The cropped T-SNE visualization of latent $\bm{z}$ from DMR.}
         \label{fig:dmr-tsne-crop}
    \end{subfigure}
    \caption{The cropped T-SNE visualization of latent $\bm{z}$ from DMR. Each $\bm{z}$ is coupled with its top 3 markers from \emph{z2m} mapping.}
    \label{fig:tsne-crop}
\end{figure}

The latent space is not interpretable, but DMR has a transition matrix that outputs a distribution of markers, which reveals what a particular dimension may encode. 

To analyze the latent space, we use $\psi_{w2}$ (Eq.~\ref{eq:softmax}) as the corresponding embedding vectors and perform T-SNE visualization of the latent $\bm{z}$, similar to what Discover~\cite{sileo2019mining} does using the softmax weight at the final prediction layer. The complete T-SNE result can be found in Appendix~\ref{supp:entire-z2c-mapping}. What we observe is an emerging hierarchical pattern, in addition to proximity. That is, while synonymous markers are clustered as expected, semantically related clusters are often closer. Fig.~\ref{fig:dmr-tsne-crop} shows the top left corner of the T-SNE result.
We can see that the temporal connectives and senses are located in the top left corner. According to their coupled markers, we can recover the semantic of these latent $\bm{z}$: preceding ($z_{27}$), succeeding ($z_{25}$, $z_{22}$, $z_{16}$) and synchronous ($z_{15}$) form nearby but separated clusters. 

For a comparison with connective-based prompting approaches, we also demonstrate the T-SNE visualization of marker representations from BASE in Fig.~\ref{fig:plm-tsne-crop}. Unlike semantically aligned vector space of DMR, locality of markers in the space of BASE representation is determined by surface form of markers and shifted from their exact meaning. Marker representations of the model w/o latent $z$ are closer because of similar lexical formats instead of underlying discourse.

From \emph{z2m} mapping, we can take a step further to analyze the correlation between markers learned by DMR. Table \ref{tab:m2z} shows the top 2 corresponding clusters of three randomly sampled markers. We can observe correlations between markers like polysemy and synonym.

\paragraph{Understanding Entanglement}
\label{sec:exp:discourse}
Labeling discourse relations is challenging since some of them can correlate, and discern the subtleties can be challenging. For example,
\emph{List} strongly correlates with \emph{Conjunction} and the two are hardly distinguishable. 

\begin{figure}[t]
    \centering
    \begin{subfigure}[t]{0.39\columnwidth}
         \centering
         \includegraphics[width=\textwidth]{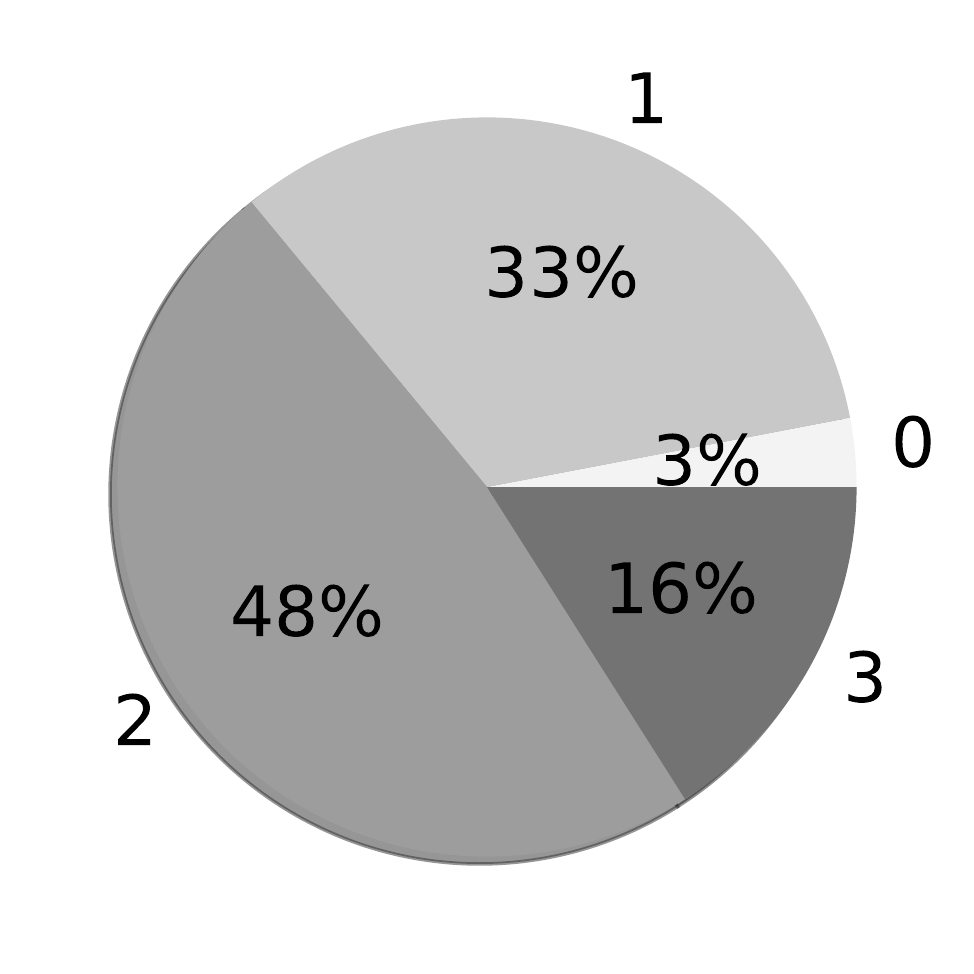}
         \caption{}
         \label{fig:rel_he:num_correct}
    \end{subfigure}
    \begin{subfigure}[t]{0.59\columnwidth}
         \centering
         \includegraphics[width=\textwidth]{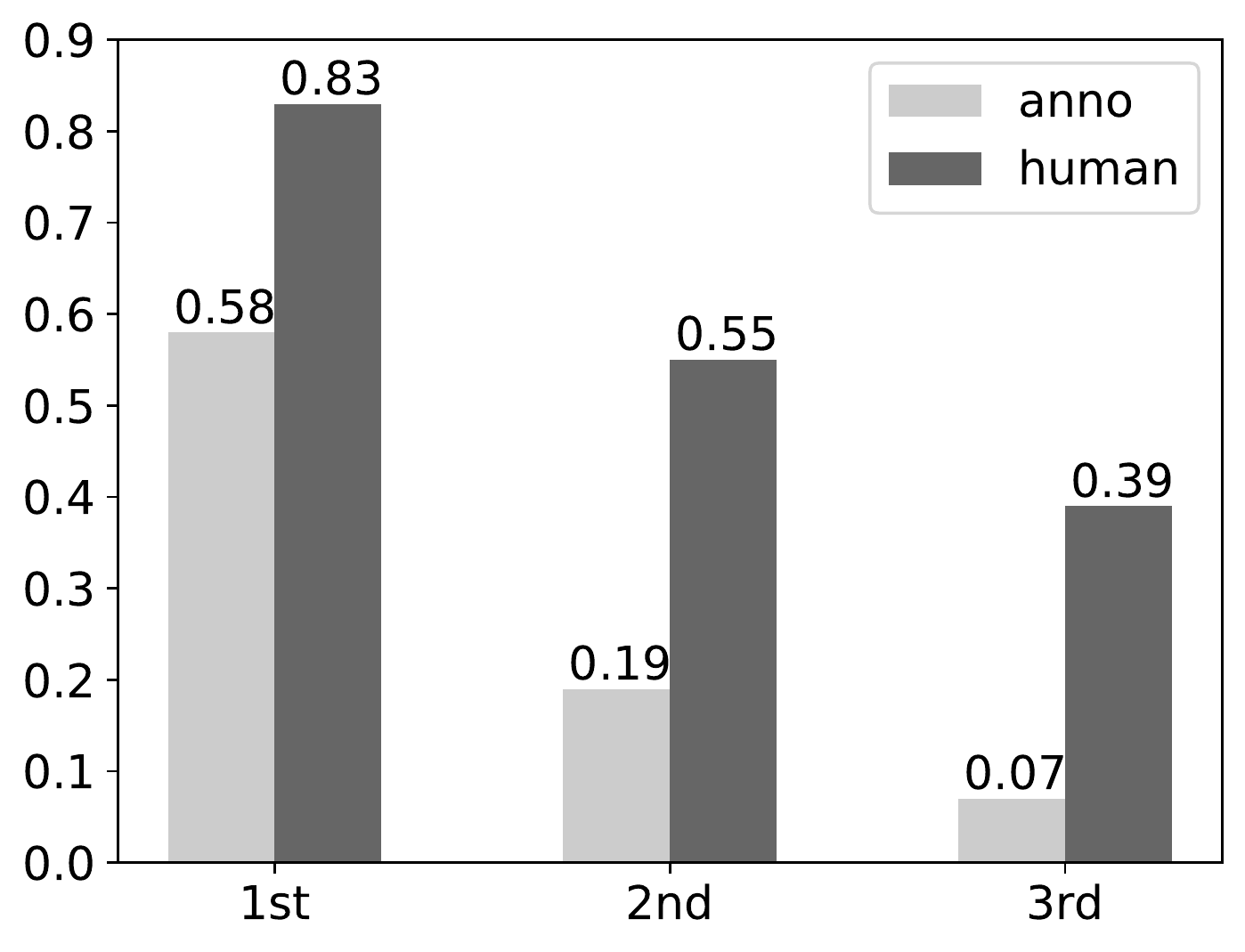}
         \caption{}
         \label{fig:rel_he:rel_acc}
    \end{subfigure}
    \caption{Human Evaluation. Figure~(a) shows numbers of reasonable relations in top-3 predictions. Figure~(b) shows the accuracy for each of the top-3 predictions evaluated by annotations or human, respectively.}
    \label{fig:rel_he}
\end{figure}

\begin{figure}
    \centering
    \includegraphics[width=\linewidth]{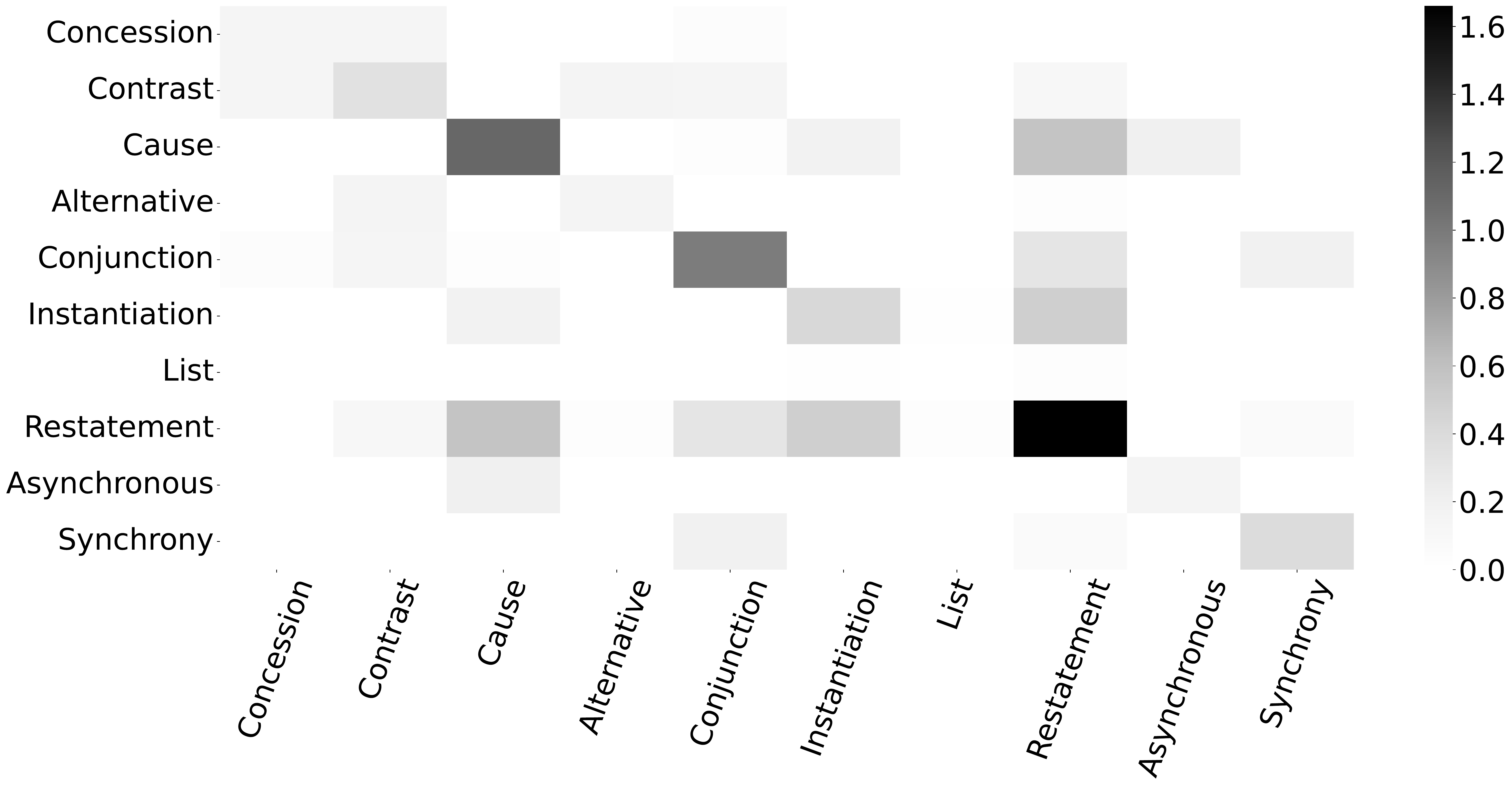}
    \caption{Confusion on Discourse Relations. We use entropy as the metric for filtering most confusing examples. We use the top-3 predictions of the 20 most confusing examples to show the entanglement between relations. We use accumulated $p(r_i)\cdot p(r_j)$ as weights for a pair of relations $r_i, r_j$. Note that implausible predictions are suppressed to ignore model errors.}
    \label{fig:rel_confusion}
\end{figure}

\begin{table*}[]
    \centering
    \scalebox{0.9}{
    \begin{tabular}{|m{5cm}|m{5cm}|c|c|c}
        \hline
        $s_1$ & $s_2$ & markers & relations \\ \hline
        Rather, they tend to have a set of two or three favorites & Sometimes, they'll choose Ragu spaghetti sauce & \makecell{because\_of\_this\\ therefore\\ for\_example\\ for\_instance} & \makecell{Contingency.Cause \\ Expansion.Instantitation} \\\hline
        
        It just makes healthy businesses subsidize unhealthy ones and gives each employer less incentive to keep his workers healthy & the HIAA is working on a proposal to establish a privately funded reinsurance mechanism to help cover small groups that can't get insurance without excluding certain employees & \makecell{because\_of\_this\\ conversely\\ therefore\\ in\_contrast} & \makecell{Contingency.Cause\\ Comparison.Contrast} \\\hline
        
        The Hart-Scott filing is then reviewed and any antitrust concerns usually met & Typically, Hart-Scott is used now to give managers of target firms early news of a bid and a chance to use regulatory review as a delaying tactic & \makecell{although \\ though \\ besides \\ also} & \makecell{Comparison.Concession \\ Expansion.Conjunction} \\
        \hline
    \end{tabular}
    }
    \caption{Case Study on Marker Ambiguity and Discourse Relation Entanglement.}
    \label{tab:cases}
\end{table*}

DMR is trained to predict a \emph{distribution} of markers, thus we expect its hidden state to capture the distribution of relations as well even when the multi-sense labels are scarce.
We drew 100 random samples and ask two researchers to check whether each of the corresponding top-3 predictions is valid and give a binary justification\footnote{The annotators achieve a substantial agreement with a Kappa coefficient of 0.68.}. Fig.~\ref{fig:rel_he:num_correct} shows that a considerable amount of 64\% examples have two or more relations evaluated as reasonable in top-3 predictions, much higher than 2.3\% multi-sense labels in PDTB2. This suggests that one way to improve upon the lack of multi-sense annotation is to use DMR to provide candidates for the annotators. For these samples, we also inspect annotator agreement in PDTB2 (Fig.~\ref{fig:rel_he:rel_acc}). While the trend is consistent with what DMR reports, it also validates again that the PTDB2 annotators under-labeled multi-senses.

To gain a deeper understanding of relation correlation, we rank the sentence pairs according to the entropy of relation prediction, a higher entropy suggests more model uncertainty, namely more confusion.

We use the top-3 predictions of the 20 highest entropy examples to demonstrate highly confusing discourse relations as shown in Fig.~\ref{fig:rel_confusion}. The accumulated joint probability of paired relations on these examples is computed as weights in the confusion matrix. The statistics meet our expectation that there exist specific patterns of confusion. For example, asynchronous relations are correlated with causal relations, while another type of temporal relations, synchronous ones are correlated with conjunction. A complete list of these high entropy examples is listed in Appendix~\ref{supp:human-eval}.

To further prove DMR can learn diverse distribution even when multi-sense labels are scarce, we also evaluate our model on the DiscoGeM~\cite{scholman-etal-2022-discogem}, where each instance is annotated by 10 crowd workers. The distribution discrepancy is evaluated with cross entropy. Our model, trained solely on majority labels, achieved a cross entropy score of 1.81 against all labels. Notably, our model outperforms the BMGF model (1.86) under the same conditions and comes close to the performance of the BMGF model trained on multiple labels (1.79)~\cite{yung-etal-2022-label}. These results highlight the strength of our model in capturing multiple senses within the data.

To conclude, while we believe explicit relation labeling is still useful, it is incomplete without also specifying a distribution. As such, DMR's $h_z$ or the distribution of markers are legitimate alternatives to model inter-sentence discourse.

\paragraph{Case Study on Specific Examples}
As a completion of the previous discussion on understanding entanglement in a macro perspective, we present a few examples in PDTB2 with markers and relations predicted by the DMR-based model. As demonstrated in Table~\ref{tab:cases}, the identification of discourse relations relies on different emphasis of semantic pairs. Taking the first case as an example, the connection between ``two or three favorities'' and ``Ragu spaghetti sauce'' indicates the \emph{Instantiation} relation while the connection between complete semantics of these two sentences results in \emph{Cause}. Thanks to the probabilistic modeling of discourse information in DMR, the cases demonstrate entanglement among relations and ambiguity of markers well.

\section{Conclusion}

In this paper, we propose the distributed marker representation for modeling discourse based on the strong correlation between discourse markers and relations. We design the probabilistic model by introducing a latent variable for discourse senses. We use the EM algorithm to effectively optimize the framework. The study on our well-trained DMR model shows that the latent-included model can offer a meaningful semantic view of markers. Such semantic view significantly improves the performance of implicit discourse relation recognition. Further analysis of our model provides a better understanding of discourse relations and markers, especially the ambiguity and entanglement issues.

\section*{Limitation \& Risks}
In this paper, we bridge the gap between discourse markers and the underlying relations. We use distributed discourse markers to express discourse more informatively. However, learning DMR requires large-scale data on markers. Although it's potentially unlimited in corpus, the distribution and types of markers may affect the performance of DMR. Besides, the current solution proposed in this paper is limited to relations between adjacent sentences.

Our model can be potentially used for natural language commonsense inference and has the potential to be a component for large-scale commonsense acquisition in a new form. Potential risks include a possible bias on collected commonsense due to the data it relies on, which may be alleviated by introducing a voting-based selection mechanism on large-scale data.

\bibliography{anthology,custom}
\bibliographystyle{acl_natbib}

\clearpage
\appendix




\begin{table}[]
    \centering
    \scalebox{0.8}{
    \begin{tabular}{ccc}\toprule
        Train & Valid & Test \\\midrule
        1566k & 174k & 174k \\ \bottomrule
    \end{tabular}}
    \caption{Statistics of Discovery Dataset}
    \label{tab:discovery_data}
\end{table}

\begin{table}[]
    \centering
    \scalebox{0.8}{
    \begin{tabular}{cccc} \toprule
        Relations & Train & Valid & Test \\ \midrule
        Comp.Concession  & 180 & 15 & 17 \\
        Comp.Contrast  & 1566 & 166 & 128 \\
        Cont.Cause  & 3227 & 281 & 269 \\
        Cont.Pragmatic Cause & 51 & 6 & 7 \\
        Expa.Alternative & 146 & 10 & 9 \\ 
        Expa.Conjunction  & 2805 & 258 & 200 \\
        Expa.Instantiation & 1061 & 106 & 118 \\   Expa.List  & 330 & 9 & 12 \\               Expa.Restatement & 2376 & 260 & 211 \\
        Temp.Async & 517 & 46 & 54 \\
        Temp.Sync & 147 & 8 & 14 \\
        Total  & 12406 & 1165 & 1039 \\ \bottomrule
    \end{tabular}}
    \caption{Statistics of PDTB2 Dataset}
    \label{tab:pdtb2_data}
\end{table}

\section{Implementation Details}
\label{supp:implementation}
We use Huggingface transformers (4.2.1) for the use of PLM backbones in our experiments.
For optimization, we optimize the overall framework according to Algorithm~\ref{alg:optimization}. We train the model on Discovery for 3 epochs with the learning rate for $\psi$ set to 3e-5 and the learning rate for $\phi$ set to 1e-2. The EM batchsize is set to 500 according to the trade-off between optimization efficiency and performance. The optimization requires around 40 hrs to converge in a Tesla-V100 GPU. For the experiments on PDTB2, we use them according to the LDC license for research purposes on discourse relation classification. The corresponding statistics of the two datasets are listed in Table~\ref{tab:discovery_data} and Table~\ref{tab:pdtb2_data}.


\begin{figure*}[h]
    \centering
    \includegraphics[width=\linewidth]{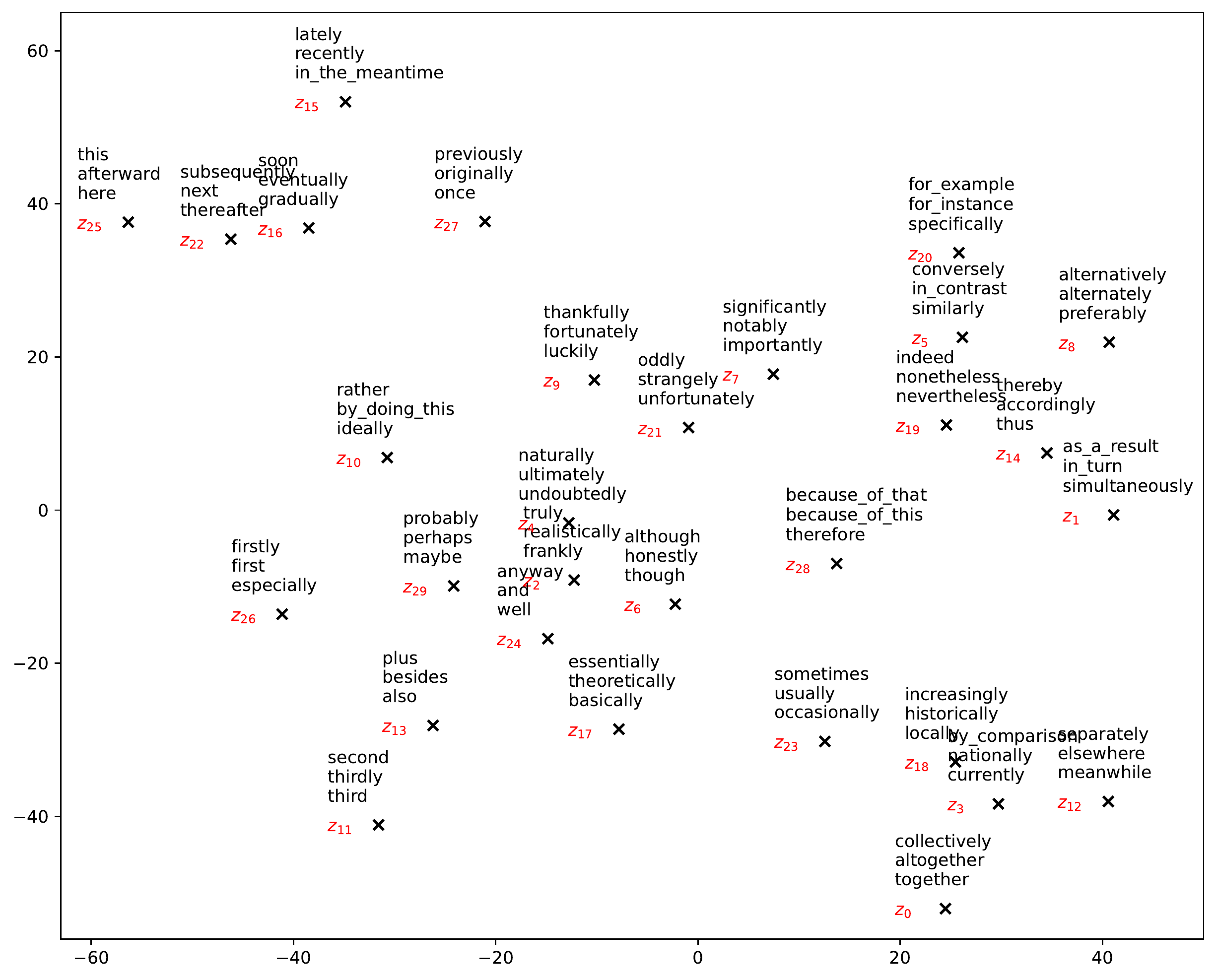}
    \caption{T-SNE Visualization of the Latent $\bm{z}$. We draw the t-sne embeddings of each latent $z$ in 2-d space with the well-trained $\psi_{w2}$ as corresponding embedding vectors. While each $z$ groups markers with similar meanings, we can also observe that related senses are clustered together. For example, temporal connectives and senses are located in the top left corner with preceding ($z_{27}$), succeeding ($z_{25}$, $z_{22}$, $z_{16}$), synchronous ($z_{15}$) ones separated. The existence of $\bm{z}$ helps to construct a hierarchical view of semantics between sentences.}
    \label{fig:tsne}
\end{figure*}

\begin{figure*}[h]
    \includegraphics[width=\linewidth]{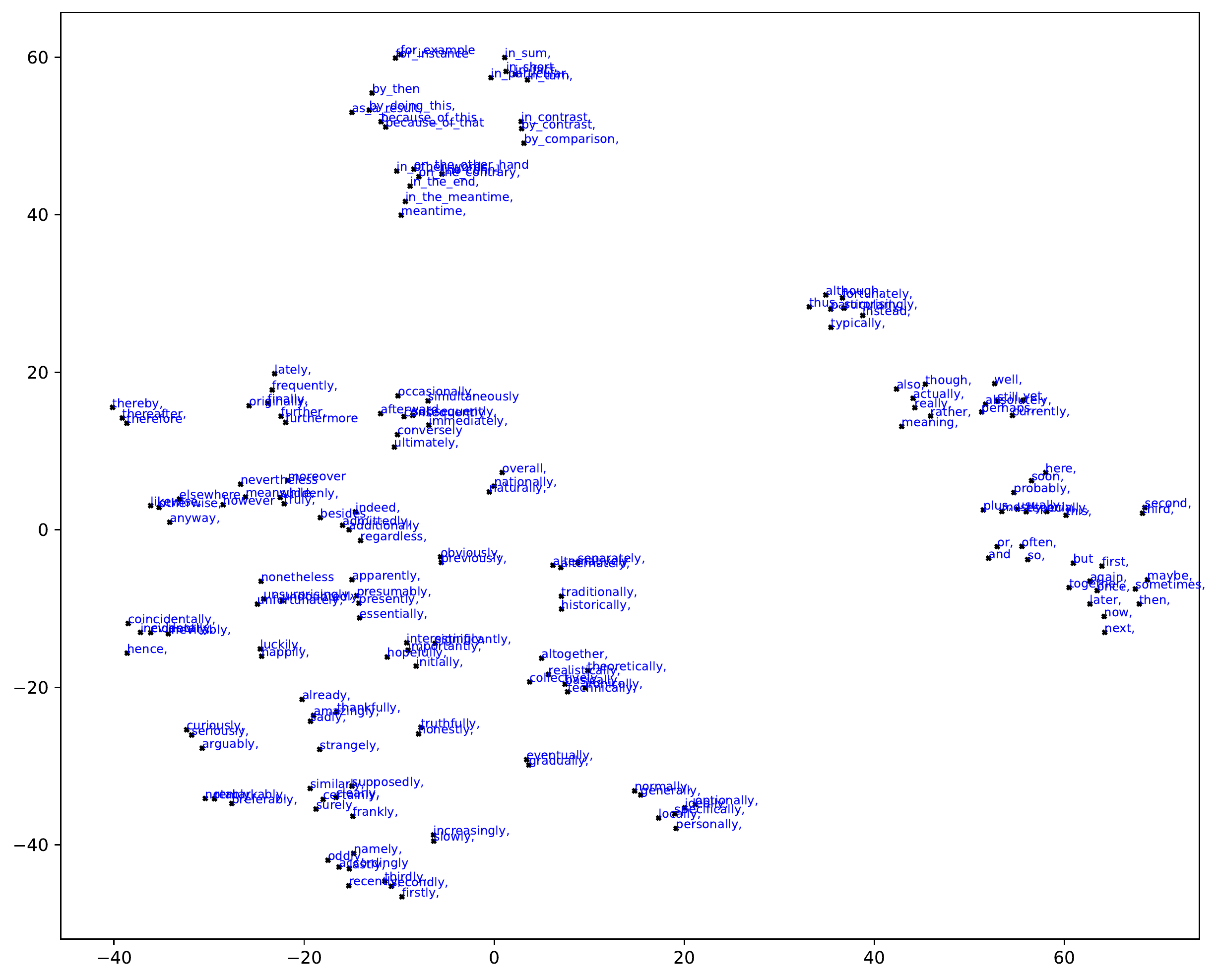}
    \caption{T-SNE Visualization of discourse markers from BASE. We draw the t-sne embeddings of each marker in 2-d space with averaged token representations of markers from BASE PLM. Comparing to the well-organized hierarchical view of latent senses in DMR, markers are not well-aligned to semantics in the representation space of BASE. It indicates the limitation of bridging markers and relations with a direct mapping.}
    \label{fig:plm-tsne}
\end{figure*}


\section{Visualization of the latent $\bm{z}$}
\label{supp:entire-z2c-mapping}
To obtain an intrinsic view of how well the connections between markers $\bm{m}$ and $\bm{z}$ can be learned in our DMR model. We draw a T-SNE 2-d visualization of $\bm{z}$'s representations in Fig.~\ref{fig:tsne} with top-3 connectives of each $z$ attached nearby. The representation vector for each $z$ is extracted from $\psi_{w2}$. The results are interesting that we can observe not only the clustering of similar connectives as $z$, but also semantically related $z$ closely located in the representation space.

\section{High Entropy Examples from Human Evaluation}
\label{supp:human-eval}

For analysis of the entanglement among relations, we did a human evaluation on randomly extracted examples from PDTB2. To better understand the entanglement among relations, we further filter the 20 most confusing examples with entropy as a metric. The entanglement is shown as Fig.\ref{fig:rel_confusion} in Sec.~\ref{sec:exp:discourse}. We list these examples in Table~\ref{tab:examples} for clarity.

\begin{table*}[ht]
    \centering
    \scalebox{0.75}{
    \begin{tabular}{p{6cm}p{6cm}ccc}
    \toprule
    $\mathbf{s_1}$ & $\mathbf{s_2}$ & \textbf{1st-pred} & \textbf{2nd-pred} & \textbf{3rd-pred} \\
    \midrule
Right away you notice the following things about a Philip Glass concert & It attracts people with funny hair & \makecell{Instantiation \\ 0.502} & \makecell{Restatement \\ 0.449} & \makecell{List \\ 0.014} \\
There is a recognizable musical style here, but not a particular performance style & The music is not especially pianistic & \makecell{\textcolor{red}{Restatement} \\ 0.603} & \makecell{Conjunction \\ 0.279} & \makecell{\textcolor{red}{Instantiation} \\ 0.048} \\
Numerous injuries were reported & Some buildings collapsed, gas and water lines ruptured and fires raged & \makecell{Restatement \\ 0.574} & \makecell{Instantiation \\ 0.250} & \makecell{List \\ 0.054} \\
this comparison ignores the intensely claustrophobic nature of Mr. Glass's music & Its supposedly austere minimalism overlays a bombast that makes one yearn for the astringency of neoclassical Stravinsky, the genuinely radical minimalism of Berg and Webern, and what in retrospect even seems like concision in Mahler & \makecell{Cause \\ 0.579} & \makecell{Restatement \\ 0.319} & \makecell{Instantiation \\ 0.061} \\
The issue exploded this year after a Federal Bureau of Investigation operation led to charges of widespread trading abuses at the Chicago Board of Trade and Chicago Mercantile Exchange & While not specifically mentioned in the FBI charges, dual trading became a focus of attempts to tighten industry regulations & \makecell{Cause \\ 0.504} & \makecell{Asynchronous \\ 0.400} & \makecell{\textcolor{red}{Conjunction} \\ 0.045} \\
A menu by phone could let you decide, `I'm interested in just the beginning of story No. 1, and I want story No. 2 in depth & You'll start to see shows where viewers program the program & \makecell{\textcolor{red}{Cause} \\ 0.634} & \makecell{\textcolor{red}{Conjunction} \\ 0.188} & \makecell{\textcolor{red}{Asynchronous} \\ 0.116} \\
His hands sit farther apart on the keyboard.Seventh chords make you feel as though he may break into a (very slow) improvisatory riff & The chords modulate & \makecell{\textcolor{red}{Cause} \\ 0.604} & \makecell{Conjunction \\ 0.266} & \makecell{\textcolor{red}{Restatement} \\ 0.082} \\
His more is always less & Far from being minimalist, the music unabatingly torments us with apparent novelties not so cleverly disguised in the simplicities of 4/4 time, octave intervals, and ragtime or gospel chord progressions & \makecell{Cause \\ 0.456} & \makecell{Restatement \\ 0.433} & \makecell{Instantiation \\ 0.052} \\
It requires that "discharges of pollutants" into the "waters of the United States" be authorized by permits that reflect the effluent limitations developed under section 301 & Whatever may be the problems with this system, it scarcely reflects "zero risk" or "zero discharge & \makecell{\textcolor{red}{Contrast} \\ 0.484} & \makecell{Cause \\ 0.387} & \makecell{\textcolor{red}{Concession} \\ 0.072} \\
The study, by the CFTC's division of economic analysis, shows that "a trade is a trade & Whether a trade is done on a dual or non-dual basis doesn't seem to have much economic impact & \makecell{Restatement \\ 0.560} & \makecell{Conjunction \\ 0.302} & \makecell{Cause \\ 0.095} \\
Currently in the middle of a four-week, 20-city tour as a solo pianist, Mr. Glass has left behind his synthesizers, equipment and collaborators in favor of going it alone & He sits down at the piano and plays & \makecell{Restatement \\ 0.357} & \makecell{Synchrony \\ 0.188} & \makecell{\textcolor{red}{Asynchronous} \\ 0.115} \\
For the nine months, Honeywell reported earnings of \$212.1 million, or \$4.92 a share, compared with earnings of \$47.9 million, or \$1.13 a share, a year earlier & Sales declined slightly to \$5.17 billion & \makecell{Conjunction \\ 0.541} & \makecell{\textcolor{red}{Contrast} \\ 0.319} & \makecell{Synchrony \\ 0.109} \\
The Bush administration is seeking an understanding with Congress to ease restrictions on U.S. involvement in foreign coups that might result in the death of a country's leader & that while Bush wouldn't alter a longstanding ban on such involvement, "there's a clarification needed" on its interpretation & \makecell{Restatement \\ 0.465} & \makecell{\textcolor{red}{Conjunction} \\ 0.403} & \makecell{Cause \\ 0.094} \\
    \bottomrule
    \end{tabular}
    }
\end{table*}

\begin{table*}[!t]
\scalebox{0.75}{
    \begin{tabular}{p{6cm}p{6cm}ccc}
\toprule
$\mathbf{s_1}$ & $\mathbf{s_2}$ & \textbf{1st-pred} & \textbf{2nd-pred} & \textbf{3rd-pred} \\
\midrule
With "Planet News Mr. Glass gets going & His hands sit farther apart on the keyboard & \makecell{Synchrony \\ 0.503} & \makecell{\textcolor{red}{Asynchronous} \\ 0.202} & \makecell{\textcolor{red}{Cause} \\ 0.147} \\
The Clean Water Act contains no "legal standard" of zero discharge & It requires that "discharges of pollutants" into the "waters of the United States" be authorized by permits that reflect the effluent limitations developed under section 301 & \makecell{Alternative \\ 0.395} & \makecell{Contrast \\ 0.386} & \makecell{Restatement \\ 0.096} \\
Libyan leader Gadhafi met with Egypt's President Mubarak, and the two officials pledged to respect each other's laws, security and stability & They stopped short of resuming diplomatic ties, severed in 1979 & \makecell{Contrast \\ 0.379} & \makecell{Concession \\ 0.373} & \makecell{Conjunction \\ 0.129} \\
His hands sit farther apart on the keyboard.Seventh chords make you feel as though he may break into a (very slow) improvisatory riff.The chords modulate, but there is little filigree even though his fingers begin to wander over more of the keys & Contrasts predictably accumulate & \makecell{Conjunction \\ 0.445} & \makecell{Synchrony \\ 0.303} & \makecell{\textcolor{red}{List} \\ 0.181} \\
NBC has been able to charge premium rates for this ad time & but to be about 40\% above regular daytime rates & \makecell{Conjunction \\ 0.409} & \makecell{Restatement \\ 0.338} & \makecell{Contrast \\ 0.224} \\
Mr. Glass looks and sounds more like a shaggy poet describing his work than a classical pianist playing a recital & The piano compositions are relentlessly tonal (therefore unthreatening), unvaryingly rhythmic (therefore soporific), and unflaggingly harmonious but unmelodic (therefore both pretty and unconventional & \makecell{Cause \\ 0.380} & \makecell{Instantiation \\ 0.323} & \makecell{Restatement \\ 0.241} \\
It attracts people with funny hair & Whoever constitute the local Left Bank come out in force, dressed in black & \makecell{\textcolor{red}{Cause} \\ 0.369} & \makecell{\textcolor{red}{Asynchronous} \\ 0.331} & \makecell{Conjunction \\ 0.260} \\
\bottomrule
    \end{tabular}
}
\caption{High Entropy Examples of Model Inference on Implicit Discourse Relation Classification}
\label{tab:examples}
\end{table*}

\end{document}